\documentclass[letterpaper, 10 pt, conference]{ieeeconf}  

\IEEEoverridecommandlockouts                              
\usepackage{xcolor}
\usepackage{amsmath}
\usepackage{makecell}
\title{\LARGE \bf
Personalised Explanations in Long-term Human-Robot Interactions
}

\author{Ferran Gebellí$^{*}$, Anaís Garrell$^{\dagger}$, Jan-Gerrit Habekost$^{\ddagger}$, Séverin Lemaignan$^{*}$, Stefan Wermter$^{\ddagger}$ and Raquel Ros$^{\S}$
\thanks{$^{*}$ Ferran Gebellí and Séverin Lemaignan belong to PAL robotics (Barcelona, Spain)
        {\tt\small ferran.gebelli@pal-robotics.com, severin.lemaignan@pal-robotics.com }}%
\thanks{$^{\dagger}$Anaís Garrell belongs to the Institut de Robòtica i Informàtica Industrial (IRI-CSIC),
and Universitat Politècnica de Catalunya - BarcelonaTech (UPC), (Barcelona, Spain)         
{\tt\small anais.garrell@upc.edu}}%
\thanks{$^{\ddagger}$Jan-Gerrit Habekost and Stefan Wermter belong to the University of Hamburg, (Hamburg, Germany)
{\tt\small jan-gerrit.habekost@uni-hamburg.de, stefan.wermter@uni-hamburg.de}}%
\thanks{$^{\S}$Raquel Ros belongs to the Artificial Intelligence Research Institute (IIIA-CSIC), (Barcelona, Spain)
        {\tt\small raquel.ros@iiia.csic.es}}%
}

\usepackage{multirow}

\begin{document}

\maketitle
\thispagestyle{empty}
\pagestyle{empty}

\begin{abstract}
In the field of Human-Robot Interaction (HRI), a fundamental challenge is to facilitate human understanding of robots. The emerging domain of eXplainable HRI (XHRI) investigates methods to generate explanations and evaluate their impact on human-robot interactions. Previous works have highlighted the need to personalise the level of detail of these explanations to enhance usability and comprehension. Our paper presents a framework designed to update and retrieve user knowledge-memory models, allowing for adapting the explanations' level of detail while referencing previously acquired concepts. Three architectures based on our proposed framework that use Large Language Models (LLMs) are evaluated in two distinct scenarios: a hospital patrolling robot and a kitchen assistant robot. Experimental results demonstrate that a two-stage architecture, which first generates an explanation and then personalises it, is the framework architecture that effectively reduces the level of detail only when there is related user knowledge. 

\end{abstract}

\section{INTRODUCTION}
\label{sec:intro}
In eXplainable Human-Robot Interaction (XHRI), explainability is acknowledged as a key factor for improving human understanding of robots' behaviours and decision-making~\cite{miller2019explanation, verhagen2021two}. From the perspective of \textit{Theory of Mind (ToM)}, XHRI has been framed as a model reconciliation problem \cite{hellstrom2018understandable, sreedharan2021foundations}.
Within these theoretical frameworks, a human is assumed to possess a mental model of the robot. In this context, the objective of explainability is to minimise the discrepancy between the robot's internal model of itself and the human's mental model of the robot, which the robot should estimate as precisely as possible.

Since robots must infer different models for each individual, personalisation becomes essential~\cite{setchi2020explainable}, but very few studies have explicitly addressed the challenge of personalisation in XHRI \cite{anjomshoae2019explainable}. Although providing detailed explanations can improve human understanding, excessive detail can lead to cognitive overload and reduced attention \cite{kulesza2012tell}. Overly complex or redundant information can confuse users \cite{mualla2020human}; therefore, explanations should be appropriate and concise \cite{mualla2022quest}, as simpler explanations are generally more comprehensible \cite{korpan2025ev}. In scenarios involving prolonged interactions, which has been identified as a common yet understudied setting in HRI \cite{belpaeme2020advice},
the need for clear explanations becomes even more critical. In such cases, personalisation is particularly important to adjust the level of detail and prevent redundant explanations of previously understood information.

\begin{figure}
\centering
\includegraphics[width=0.49\textwidth]{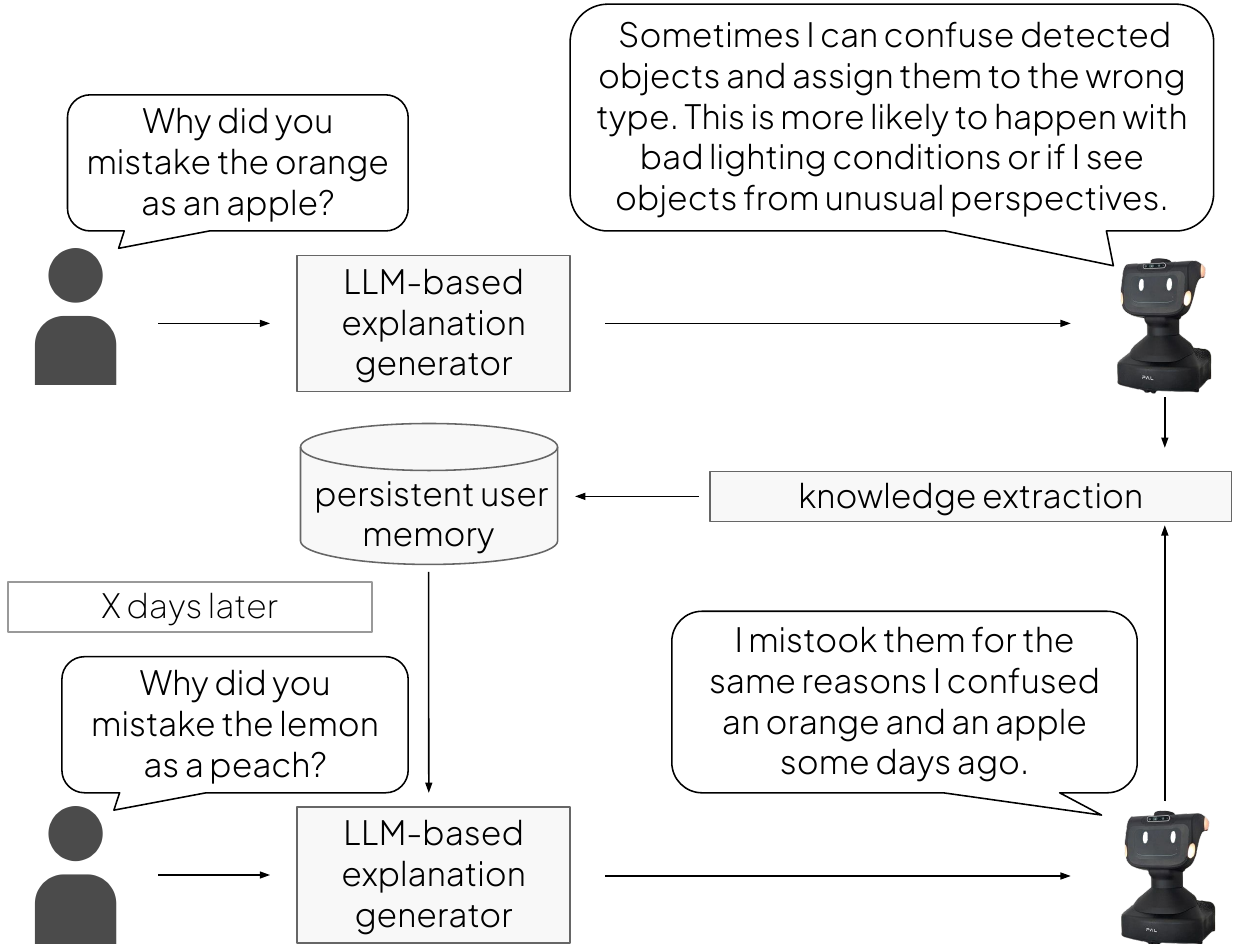}
\caption{Illustrative personalisation of an explanation's level of detail in a second interaction after some days elapsed since the first interaction.}
\vspace{-0.5em}
\label{fig:overview_architecture}
\end{figure}

In this study, we propose a framework that uses Large Language Models (LLMs) to generate personalised explanations for end users, as illustrated in Fig. \ref{fig:overview_architecture}. Our key contribution is the integration and retrieval of an external, user-specific memory that is continuously updated to track the user's estimated knowledge. We validate the effectiveness of the framework concerning the explanation's level of detail while comparing three architectures for querying and retrieving user-specific known concepts\footnote{The implementation code and experiments are available at https://github.com/fgebelli/personalised\_XHRI}. Future research will focus on evaluating the impact of these personalised explanations on user usability. We determine two research questions:
\begin{itemize}
    \item \textbf{RQ1}: Can LLMs provide explanations about robot behaviours and decision-making that provide an adapted level of detail?
    \item \textbf{RQ2}: Which knowledge retrieval strategies are most effective in producing robust personalised explanations?
\end{itemize}

Our paper is organised as follows. Sec. \ref{sec:related} reviews the related literature. Sec. \ref{sec:approach} details the proposed approach. Then, Sec. \ref{sec:exp} details the experiments' procedure, while Sec. \ref{sec:results} presents the results. A discussion of the results is provided in Sec. \ref{sec:discussion}. Finally, Sec. \ref{sec:conclusions} concludes the paper.

\section{RELATED WORK}
\label{sec:related}

In HRI, user modelling and personalisation have been extensively studied~\cite{rossi2017user}, primarily to adapt decision-making processes and behaviours. This has been approached through classical methods~\cite{andriella2019learning} and, more recently, with LLMs~\cite{wu2023tidybot}.
However, personalisation in HRI has been largely overlooked in adapting dialogues to explain the reasoning behind these decisions and behaviours.

\subsection{Users' knowledge tracing}

A related area to personalised explanations is intelligent tutoring systems, which track users' knowledge. Various approaches have been proposed, including Learning Factors Analysis~\cite{cen2006learning}, Bayesian Knowledge Tracing~\cite{corbett1994knowledge}, and Deep Knowledge Tracing~\cite{piech2015deep}, along with numerous extensions.

Nevertheless, these methods typically assume the availability of abundant and structured user feedback to update the learning process. In HRI scenarios, it is often impractical to solicit frequent feedback from users as it interrupts interactions, posing a significant challenge to direct application.

\subsection{Personalised XAI}
In non-embodied eXplainable Artificial Intelligence (XAI), the personalisation of explanations has been explored from multiple perspectives. In~\cite{soni2021not}, explanations are tailored to different types of users, which are identified using clustering techniques. Other studies focus on individual-level personalisation rather than general user categories~\cite{verhagen2023personalized}. Personalisation can also be based on user preferences~\cite{boggess2020towards} or a combination of preferences and performance~\cite{silva2024towards}.

However, these studies focus on non-embodied systems that do not encounter some of the complexities inherent to robots. In robotic systems, explanations are intrinsically more complex due to the robot’s ability to physically interact with the environment, which extends the behaviours and decision-making to be explained. Additionally, the anthropomorphic attributes ascribed to robots influence human beliefs and expectations in ways that differ from interactions with purely digital systems, pushing the need for explainability.

\subsection{Personalised LLMs}
LLMs are a promising tool for personalising conversations. In addition to tracking different chat histories for individual users, more advanced techniques have been proposed. Some approaches leverage user context through prompting mechanisms~\cite{wu2023tidybot}, though these methods are not well-suited for handling extensive historical interactions. To address this limitation, alternative strategies employ Retrieval-Augmented Generation (RAG) to extract relevant excerpts from previous conversations~\cite{salemi2024optimization, richardson2023integrating}. Another widely adopted approach involves fine-tuning LLMs based on past user interactions~\cite{wozniak2024personalized, tan2024personalized} or constructing user-specific embeddings~\cite{liu2024llms+}. Furthermore, other works have studied explicit external memories that are used to store corrections that users provide about the system's knowledge~\cite{madaan2022memory} or user-specified preferences~\cite{joko2024doing}.

Personalised LLMs have also been investigated in the context of tabletop robots. In~\cite{irfan2023between}, an LLM personalises interactions by incorporating summaries of prior open-domain conversations with older adults. Similarly,~\cite{onorati2023creating} explores adapted conversations based on the social network accounts followed by each user.

Although these studies provide personalisation, they do not incorporate complex associations to the embodiment capabilities nor an explainability perspective to explicitly model the knowledge that users have about robots.

\subsection{LLM-based explanations in robots}
LLM-based explanations have been proposed in robotics, yet they lack the personalisation component. Template-based approaches have been presented~\cite{lemasurier2024templated}, where the robot’s behaviour trees are integrated into prompts to generate explanations. Another approach utilises RAG to leverage the robot’s logs~\cite{sobrin2024explaining}, addressing the limitations of overly long context windows. Additionally, other studies propose the generation of episodic memories based on the automatic detection of key events, capturing both the robot’s internal states and relevant environmental information~\cite{liu2023reflect, wangcan}.

However, these studies do not incorporate the personalisation perspective, an unaddressed issue not only in LLM-based explanation methods but more broadly in XAI~\cite{anjomshoae2019explainable}.

Therefore, we have introduced in our previous work~\cite{gebelli2025pe} an initial framework for personalised LLM-based explanations, drawing inspiration from works such as~\cite{sobrin2024explaining, liu2023reflect, wangcan}, while adding the personalisation component. In the present work, we further refine this framework and conduct extensive validation of the adapted explanations' level of detail.

\section{A FRAMEWORK FOR PERSONALISED XHRI}
\label{sec:approach}

In this section, we introduce a framework for personalised explanations in HRI that tracks and updates individual memory models for interacting users\footnote{We assume that an application-specific and privacy-aware system provides the user re-identification that selects the adequate memory model.}.
In our framework, LLMs provide personalised explanations based on relevant \textit{user context} and \textit{robot context} that are included in the prompts.

Computing the \textit{user context} is central within the framework, as it is directly relevant to personalisation. We present 4 key modules in our framework: (1) a \textit{persistent user memory} keeps track of a discrete list of concepts that constitute the user's knowledge about the robot, (2) a \textit{knowledge extraction} module populates the \textit{persistent user memory} from previous explanations, (3) a \textit{retrieval of relevant user knowledge} exploits embedding similarity between the user question and previously known concepts in the \textit{persistent user memory}, and (4) a \textit{post-processing} step transforms each of the retrieved concepts to compute the \textit{user context}.

Since the primary objective of our work is personalisation rather than complex explanation generation, our framework assumes that the relevant robot internal information to generate the explanation is available within the \textit{robot context}, following available approaches in the literature, e.g.,  \cite{sobrin2024explaining, liu2023reflect, wangcan}. In this work, we employ a straightforward \textit{relevant info extraction} of recent high-level logs that provide the \textit{robot context}, which is a \textit{string} containing only the root causes and circumstances needed for the explanation generation.

In the next subsections, we first present 3 different architectures that implement our framework, and then we provide a detailed description of the core framework modules.

\begin{figure}[t]
\centering
\includegraphics[width=0.48\textwidth]{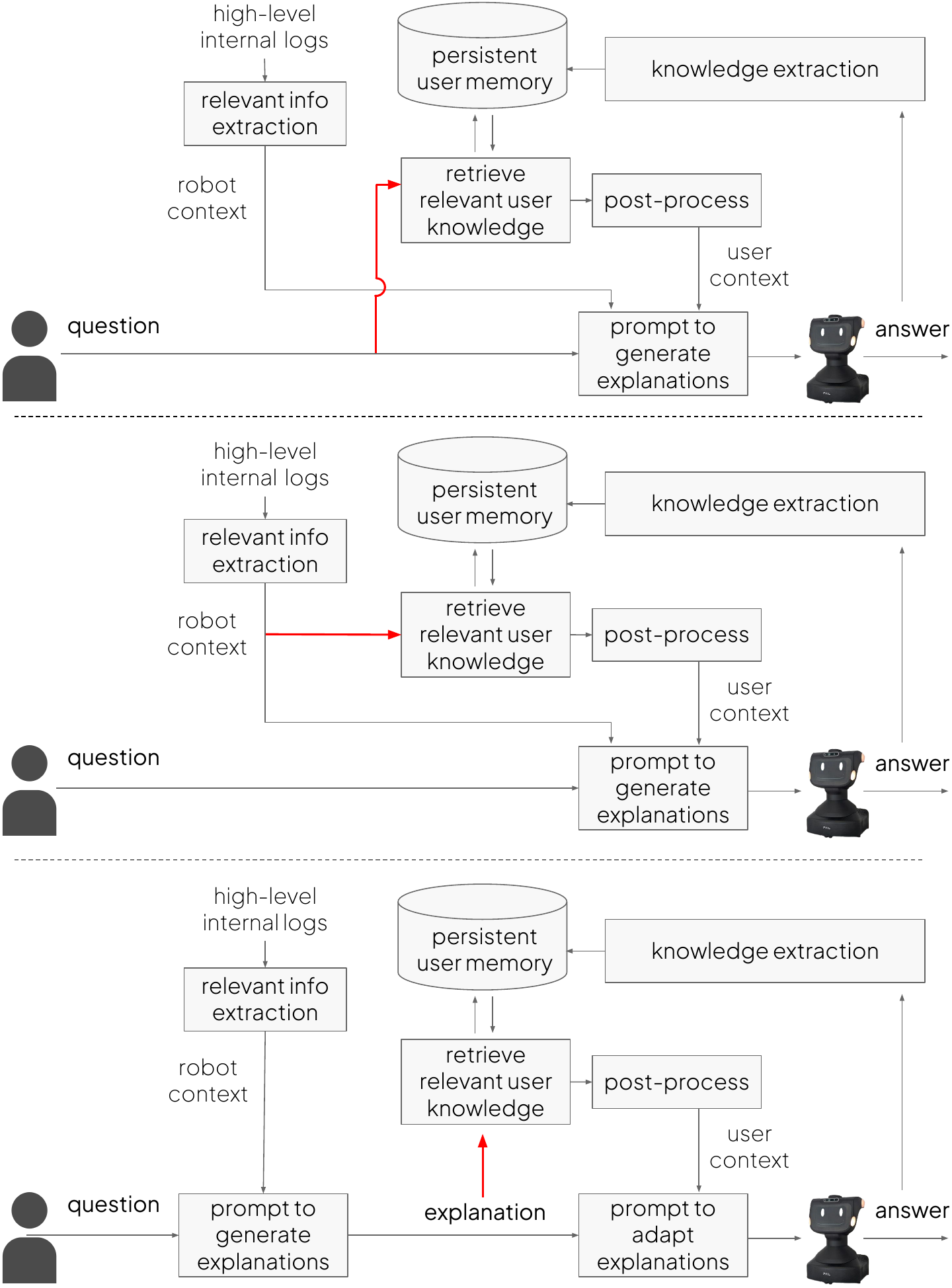}
\caption{Overview of the proposed architectures that employ either the user \textit{question query} (top), \textit{robot context query} (mid) or full \textit{explanation query} (bottom) to retrieve the user memory. Red arrows highlight the main differences.}
\vspace{-1.2em}
\label{fig:architecture}
\end{figure}

\subsection{Architecture variants}

We propose 3 different architectures in Fig.~\ref{fig:architecture}. Since our focus is on the representation and retrieval of the \textit{persistent user memory}, we aim to compare different ways to retrieve that memory to obtain the user-related knowledge, which will be later used in the personalisation of the explanation. The \textit{question query} architecture (top in Fig.~\ref{fig:architecture}) uses the user question to retrieve related concepts. For the \textit{robot context query} architecture (middle in Fig.~\ref{fig:architecture}), the \textit{robot context} is employed to retrieve the memory. Finally, the \textit{explanation query} architecture (bottom in Fig.~\ref{fig:architecture}) represents a two-stage process, in which the first stage generates a full explanation, which is used as the \textit{persistent user memory}'s query to later adapt the initial explanation in a second LLM prompt. 

We hypothesise that the query employed for the retrieval of the \textit{persistent user memory}, which we detail in the next subsection and later compare in Sec. \ref{sec:results}, will be significantly influenced by the content of the query. We expect that when the query content is semantically closer to the \textit{persistent user memory} entries, the retrieval process will be more accurate. In this direction, we expect the \textit{explanation query} architecture to perform the best, since the query content is expected to have a higher similarity to previous memories, when compared to the \textit{question query} or \textit{robot context query}. A further advantage of the \textit{explanation query} architecture is its modular design, enabling integration with any other explanation-generating systems, including those not LLM-based, to add the personalisation component. However, a disadvantage of the two-stage architecture is a potential increase in inference time. We discuss this trade-off later in Sec. \ref{sec:discussion}.

\subsection{Persistent User Memory and Knowledge Extraction}

Each user's memory model consists of a dynamic list of concepts that the user has previously acquired regarding the robot's functionality and behaviour. These concepts are \textit{strings} that capture key insights, such as ``\textit{The robot cannot reach all objects on the table}'' or ``\textit{The robot can be obstructed by obstacles}''.

Each \textit{string} concept is stored with two metadata fields: a timestamp indicating the most recent update of the concept and an embedding vector generated using the \textit{text-embedding-3-small} model from \textit{OpenAI}. When a new concept is added to the memory, its embedding vector is compared against existing entries using cosine similarity. If the similarity exceeds a predefined threshold, the new concept replaces the existing one since they essentially refer to the same user knowledge with a different rephrasing.

\begin{figure}[h]
\centering
\includegraphics[width=0.49\textwidth]{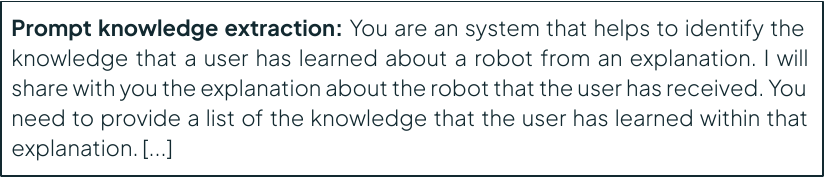}
\caption{Example prompt to extract the concepts from new conversations.}
\vspace{-0.5em}
\label{fig:prompt_extract}
\end{figure}

The \textit{persistent user memory} is populated using concepts extracted from previous explanations by a dedicated LLM (example prompt in Fig.~\ref{fig:prompt_extract}).

\subsection{Retrieval of Relevant User Knowledge}

Following the RAG paradigm, the system retrieves concepts from the \textit{persistent user memory} based on cosine similarity. Unlike traditional RAG implementations, which retrieve the top-$K$ similar entries, we introduce a similarity threshold ($\tau_{retr}$), since no user context should be retrieved in case there are no related known concepts. Filtering out unrelated prior knowledge contributes to both scalability and effectiveness, ensuring that the system incorporates in \textit{user context} only truly relevant concepts. The $\tau_{retr}$ threshold can be adjusted in a trade-off that we discuss later in Sec. \ref{sec:discussion}. 

\subsection{Post-processing of Retrieved Knowledge}

Our approach explicitly models the natural decline of knowledge due to lack of use over long-term interactions. Once the relevant concepts have been retrieved, we update the probability that the user retains each concept by applying Ebbinghaus’ exponential forgetting curve \cite{ebbinghaus1885gedachtnis}, $P = e^{-\lambda t}$, where $t$ is the time elapsed from the timestamp in the concept metadata and $\lambda$ is a decay parameter that can be adjusted based on the user type, such as young or elderly adults. This probability is post-processed to finally obtain the \textit{user context}, which we split into two components that are later introduced in the exemplified prompts in Fig.~\ref{fig:prompt_one_stage} and~\ref{fig:prompt_two_stage}:

\begin{itemize}
    \item \textit{user\_processed\_knowledge}: First, the probability values are converted into qualitative descriptors inspired by \cite{hillson2005describing}, e.g., translating probabilities above $0.9$ to \textit{very likely}. The elapsed time is also represented in a textual descriptor such as \textit{a few days ago} or \textit{about a month ago}. Each retrieved concept is then summarised in the following template:  
    \textit{``The user \{probability descriptor\} knows that \{concept\} from a conversation \{elapsed time descriptor\}.''}
    
    \item \textit{detail\_level\_instructions}:  The probabilities are weighted by the retrieval similarity scores, and the obtained values are thresholded into one of the following instructions that specify detail level target: (1) \textit{Provide a complete explanation with all the details}, (2) \textit{The explanation should be moderately short}, (3) \textit{The explanation should be concise, consisting of one or two sentences} or (4) \textit{Provide a very short summary in a single sentence}.
\end{itemize}

\subsection{Explanation Generation}

\begin{figure}
\centering
\includegraphics[width=0.49\textwidth]{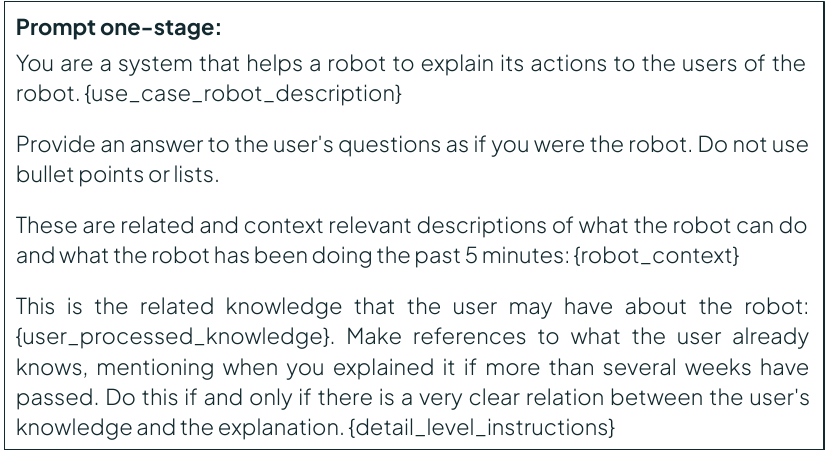}
\caption{Prompt structure for the one-stage explanation generation process.}
\label{fig:prompt_one_stage}
\vspace{-0.8em}
\end{figure}

\begin{figure}
\centering
\includegraphics[width=0.49\textwidth]{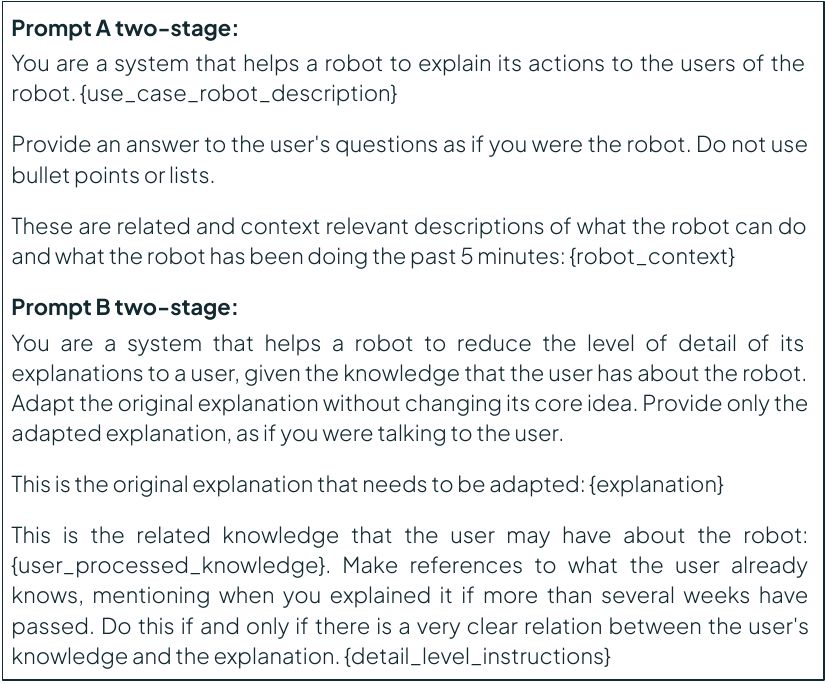}
\caption{Prompt structure for the two-stage explanation generation process.}
\label{fig:prompt_two_stage}
\vspace{-0.8em}
\end{figure}

In the \textit{question} and \textit{robot context query} architectures, a single LLM simultaneously processes the \textit{user context} and \textit{robot context} (Fig. \ref{fig:prompt_one_stage}). Conversely, in the \textit{explanation query} architecture, a first LLM generates an initial explanation based on the \textit{robot context}, while a second LLM refines this explanation to align with the user’s knowledge (Fig.~\ref{fig:prompt_two_stage}). Importantly, the second step is only executed if there are retrieved concepts; otherwise, the full explanation is forwarded.

\section{EXPERIMENTS}
\label{sec:exp}

Our approach is evaluated in two distinct use cases: a hospital patrolling robot and a kitchen assistant robot.
Inspired by real-world robot tests, we have identified a set of 7 topics that relate to knowledge related to robot events, failures and capabilities that users seek to gain. For each topic, we prepare a pool of 3 user questions.
The evaluation presented here is conducted based on synthetic data, enabling controlled interactions and a systematic analysis.

\subsection{Kitchen assistant robot use case}

\begin{figure}[h]
\centering
\includegraphics[width=0.42\textwidth]{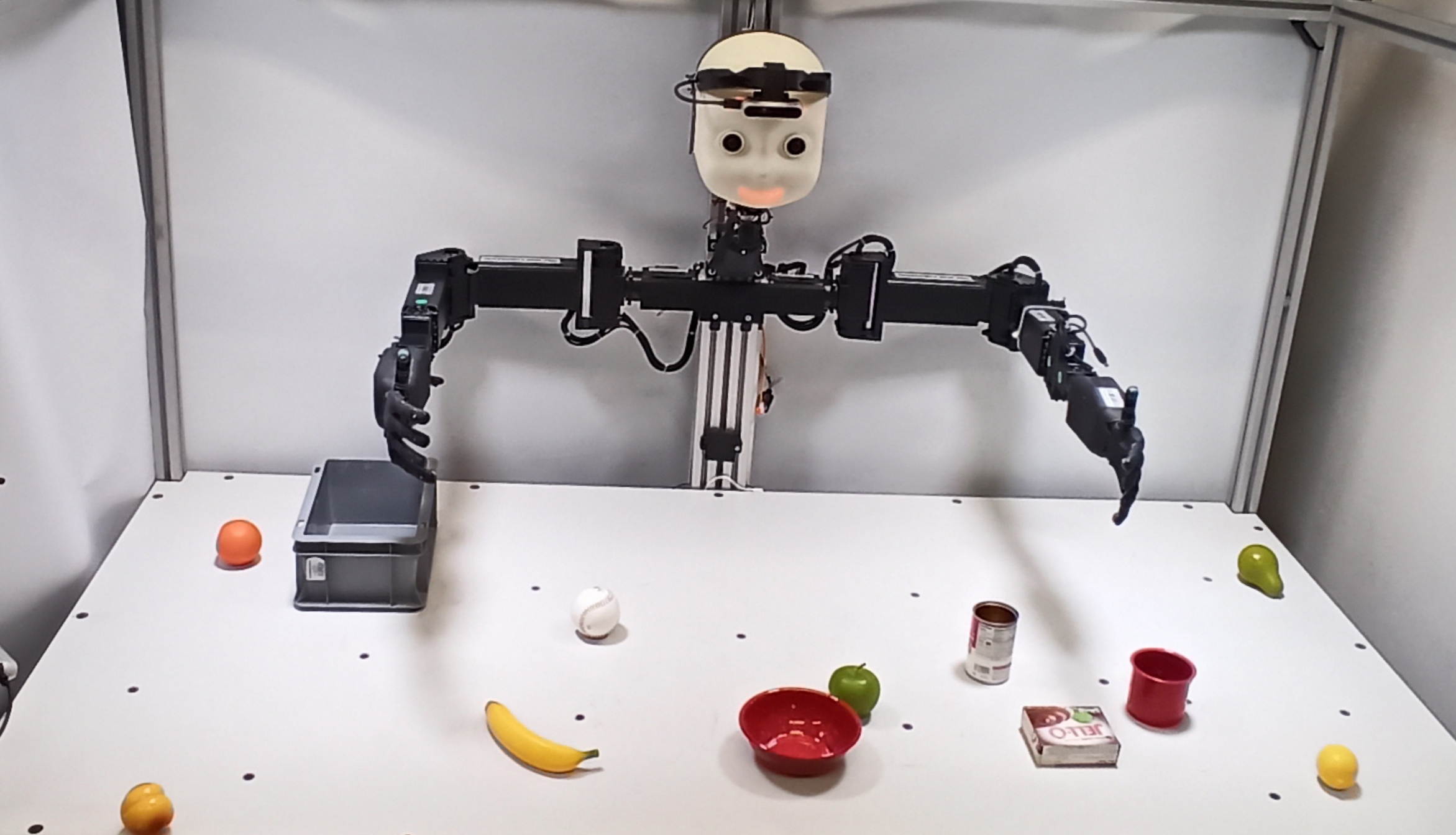}
\caption{Use case of a kitchen assistant robot.}
\vspace{-0.5em}
\label{fig:nicol}
\end{figure}

\begin{figure}[h]
\centering
\includegraphics[width=0.49\textwidth]{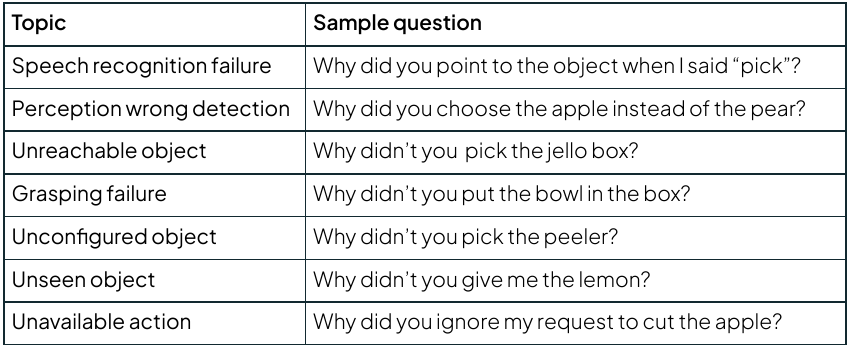}
\caption{Topics and sample questions for the kitchen assistant use case.}
\label{fig:topic_nicol}
\end{figure}

The kitchen assistant robot helps by placing objects in boxes, handing them to the user or pointing to them. Several kitchen-related objects are used, like cans bowls or fruits. 
We employ the Neuro-Inspired COLlaborator (NICOL) robot from the University of Hamburg (Fig. \ref{fig:nicol}). It is a robotic platform with a humanoid head and two 8 DoF arms that can verbally interact with users. Fig. \ref{fig:topic_nicol} details the seven topics identified for this use case, along with an example question for each one.

\subsection{Hospital patrolling robot use case}

\begin{figure}[h]
\centering
\includegraphics[width=0.42\textwidth]{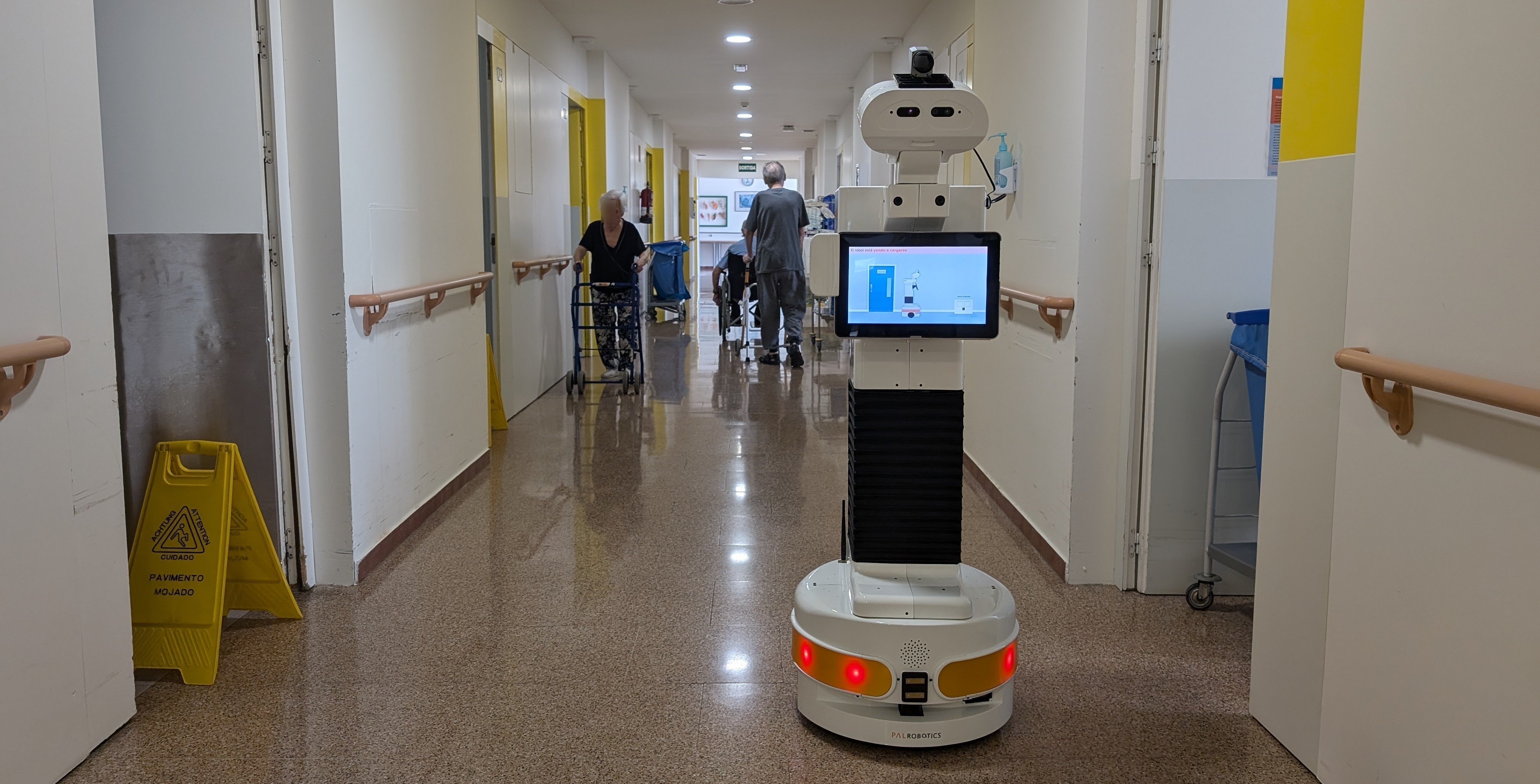}
\caption{Use case of a robot patrolling a healthcare facility.}
\vspace{-1.0em}
\label{fig:patrolling}
\end{figure}

\begin{figure}[h]
\centering
\includegraphics[width=0.49\textwidth]{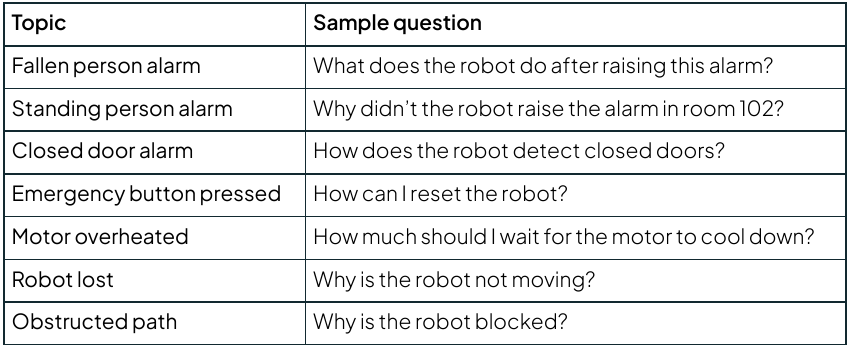}
\vspace{-1.2em}
\caption{Topics and sample questions for the hospital robot use case.}
\vspace{-0.5em}
\label{fig:topic_safely}
\end{figure}

In this use case, a robot is in charge of autonomously patrolling the rehabilitation unit of an intermediate healthcare facility. The robot is tasked to alert the nursing staff in case of hazardous situations for the patients, e.g., a fallen patient on the floor, as well as for internal failures. The robot-triggered alerts are received on all staff's phones, with an optional interface to obtain explanations. We use PAL’s Tiago LITE mobile robot (Fig. \ref{fig:patrolling}). Fig. \ref{fig:topic_safely} lists the topics and sample questions for this use case.

\subsection{Procedure}

\begin{figure}[t]
\centering
\includegraphics[width=0.44\textwidth]{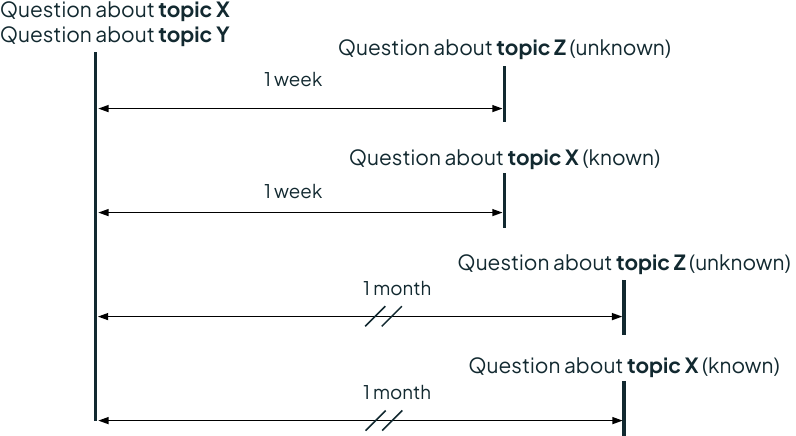}
\caption{Timeline for the 4 test cases in the experiment.}
\label{fig:timeline}
\end{figure}

We define 4 test cases, which are presented in Fig.~\ref{fig:timeline}. The motivation behind them is two-fold. On the one hand, we pursue to verify that the reduction of the level of detail is higher for recently learned concepts than for concepts that were learned longer ago, since they are less fresh in the user's memory. On the other hand, we aim to validate that the system does not produce summarised explanations in cases where the user does not hold any related knowledge, even if the user already knows concepts from unrelated topics.

Each test case repetition consists of two interactions between a user and a robot, as in Fig. \ref{fig:overview_architecture}. In the first interaction, we simulate 2 user questions given 2 hypothetical events related to 2 topics from Fig. \ref{fig:topic_nicol} or \ref{fig:topic_safely}. After a given \textit{elapsed time = \{1 week, 1 month\}} we simulate a second interaction which takes into account the first one. In this case, the questioned topic is either a known one (from the first interaction) or a completely new one, i.e. a topic never asked before (\textit{topic type = \{known, unknown\}}). The topics \textit{X, Y, Z} in Fig. \ref{fig:timeline} are always different among them. We evaluate only the explanation generated in the second interaction.

Each test case is repeated 70 times per robot use case, randomly selecting the topics and questions from the pool.

We repeat the same procedure for each of the architectural variants in Fig. \ref{fig:architecture}, which have been extended with four levels of the retrieval threshold to analyse different degrees of retrieval strictness: zero ($\tau_{retr} = 0$), low ($\tau_{retr} = 0.35$), mid ($\tau_{retr} = 0.45$) and high ($\tau_{retr} = 0.55$). For the $\tau_{retr} = 0$ the entire memory is retrieved, thus, the query content becomes irrelevant, and only the modules' distribution and connections matter. Therefore, the \textit{question query} and \textit{robot query} have been merged into a single ablation for $\tau_{retr} = 0$.

All LLMs utilised within the proposed architecture employ the \textit{OpenAI 4o-mini} model with a temperature of $0.3$.

\section{RESULTS}
\label{sec:results}

\begin{figure*}[ht]
\centering
\includegraphics[width=0.995\textwidth]{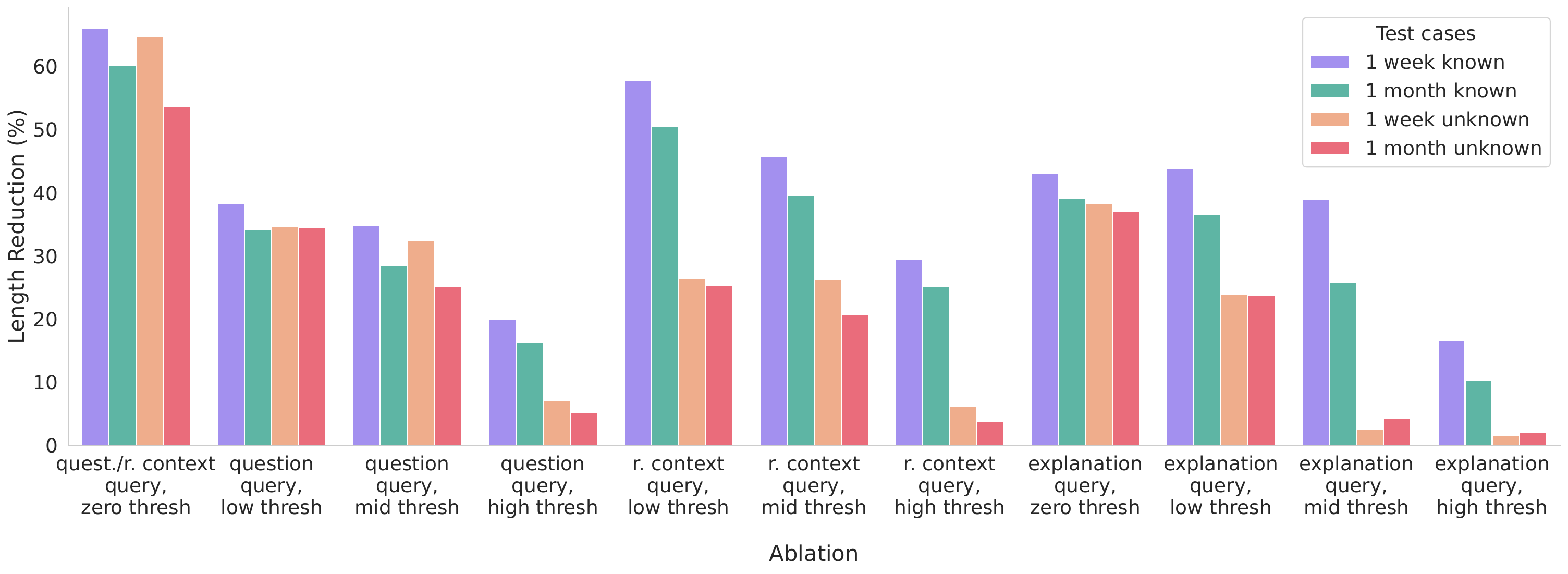}
\vspace{-1.0em}
\caption{Length reduction of the personalised explanations compared to the baseline explanation length.}
\vspace{-1.0em}
\label{fig:length_reduction}
\end{figure*}

The consolidated results are detailed in Table \ref{tab:results}, where each row is the average across 140 runs (\textit{2 use cases x 70 repetitions}). We have computed separate metrics for the retrieval process and the final adapted explanation. Regarding the retrieval process, we compute the precision, recall and false positive rate based on the ground-truth topics associated with each retrieved concept. Concerning the final response, we compute the explanation length reduction compared to a baseline explanation generated with a blank user memory for the same event, topic and question. As a reference, the average baseline explanation length is 509 characters. Fig. \ref{fig:length_reduction} plots the length reduction values.

\begin{table*}[h]
\centering
\begin{tabular}{p{0.9cm}p{1cm}|p{3.5cm}p{0.75cm}|p{1.5cm}p{1.5cm}p{2.1cm}p{2.1cm}}
\hline
\multicolumn{2}{c|}{\textbf{Test case}} & \multicolumn{2}{c|}{\textbf{Ablation}} & \multicolumn{4}{c}{\textbf{Results}}\\
\hline
\makecell[l]{\textbf{Type of} \\ \textbf{Topic}} &
\makecell[l]{\textbf{Elapsed} \\ \textbf{Time}} &
\textbf{Architecture} &
\makecell[l]{\textbf{Thresh} \\ ($\tau_{retr}$)} &
\makecell[l]{\textbf{Retrieval} \\ \textbf{Precision} \\ \textbf{(\%)}} &
\makecell[l]{\textbf{Retrieval} \\ \textbf{Recall} \\ \textbf{(\%)}} &
\makecell[l]{\textbf{Retrieval} \\ \textbf{False Positive Rate} \\ \textbf{(\%)}} &
\makecell[l]{\textbf{Response} \\ \textbf{Length Reduction} \\ \textbf{(\%)}} \\
\hline
known & 1 week & question / robot context query& zero & 51.67 & 100.00 & 48.33 & 65.97\\
&&question query& low & 54.61 & 50.36 & 22.62 & 38.33\\
&&question query& mid & 73.44 & 40.00 & 10.18 & 34.75\\
&&question query& high & 88.88 & 15.00 & 1.07 & 20.04\\
&&robot context query& low & 77.83 & 68.21 & 16.79 & 57.79\\
&&robot context query& mid & 83.14 & 45.71 & 7.86 & 45.72\\
&&robot context query& high & 92.28 & 25.00 & 1.96 & 29.53\\
&&explanation query & zero & 52.29 & 100.00 & 47.14 & 43.14\\
&&explanation query & low & 85.22 & 82.5 & 11.25 & 43.89\\
&&\textbf{explanation query} & \textbf{mid} & \textbf{94.27} & \textbf{49.64} & \textbf{2.32} & \textbf{38.98}\\
&&explanation query & high & 96.42 & 18.57 & 0.36 & 16.64\\
\hline
unknown & 1 week &question / robot context query & zero & - & - & 100.00 & 64.71\\
&&question query& low & - & - & 44.11 & 34.68\\
&&question query& mid & - & - & 24.29 & 32.38\\
&&question query& high & - & - & 3.27 & 7.03\\
&&robot context query& low & - & - & 32.68 & 26.45\\
&&robot context query& mid & - & - & 17.68 & 26.18\\
&&robot context query& high & - & - & 3.04 & 6.21\\
&&explanation query & zero & - & - & 100.00 & 38.35\\
&&explanation query & low & - & - & 19.88 & 23.86\\
&&\textbf{explanation query} & \textbf{mid} & - & - & \textbf{1.79} & \textbf{2.51}\\
&&explanation query & high & - & - & 0.36 & 1.63\\
\hline
known & 1 month &question / robot context query & zero & 51.67 & 100.00 & 48.33 & 60.17\\
&&question query & low & 56.17 & 53.21 & 23.51 & 34.17\\
&&question query& mid & 70.15 & 36.79 & 10.54 & 28.55\\
&&question query & high & 80.00 & 15.00 & 1.96 & 16.30\\
&&robot context query & low & 76.49 & 67.86 & 17.14 & 50.48\\
&&robot context query & mid & 83.33 & 45.36 & 7.68 & 39.57\\
&&robot context query& high & 92.94 & 25.36 & 1.96 & 25.22\\
&&explanation query & zero & 52.86 & 100.00 & 47.14 & 39.04\\
&&explanation query & low & 84.37 & 80.71 & 11.37 & 36.52\\
&&\textbf{explanation query }& \textbf{mid} & \textbf{96.15} & \textbf{48.57} & \textbf{1.25} & \textbf{25.84}\\
&&explanation query & high & 98.53 & 16.07 & 0.18 & 10.31\\
\hline
unknown & 1 month & question / robot context query & zero & - & - & 100.00 & 53.64\\
&&question query & low & - & - & 45.71 & 34.51\\
&&question query & mid & - & - & 22.50 & 25.19\\
&&question query & high & - & - & 2.68 & 5.27\\
&&robot context query & low & - & - & 33.21 & 25.41\\
&&robot context query & mid & - & - & 18.39 & 20.73\\
&&robot context query & high & - & - & 2.68 & 3.87\\
&&explanation query & zero & - & - & 100.00 &  37.03\\
&&explanation query & low & - & - & 20.78 & 23.81\\
&&\textbf{explanation query} & \textbf{mid} & - & - & \textbf{2.68} & \textbf{4.25}\\
&&explanation query & high & - & - & 0.54 & 2.00\\
\hline
\end{tabular}
\caption{Results for the different test cases and ablations.}
\vspace{-1.5em}
\label{tab:results}
\end{table*}

\begin{figure}
\centering
\includegraphics[width=0.46\textwidth]{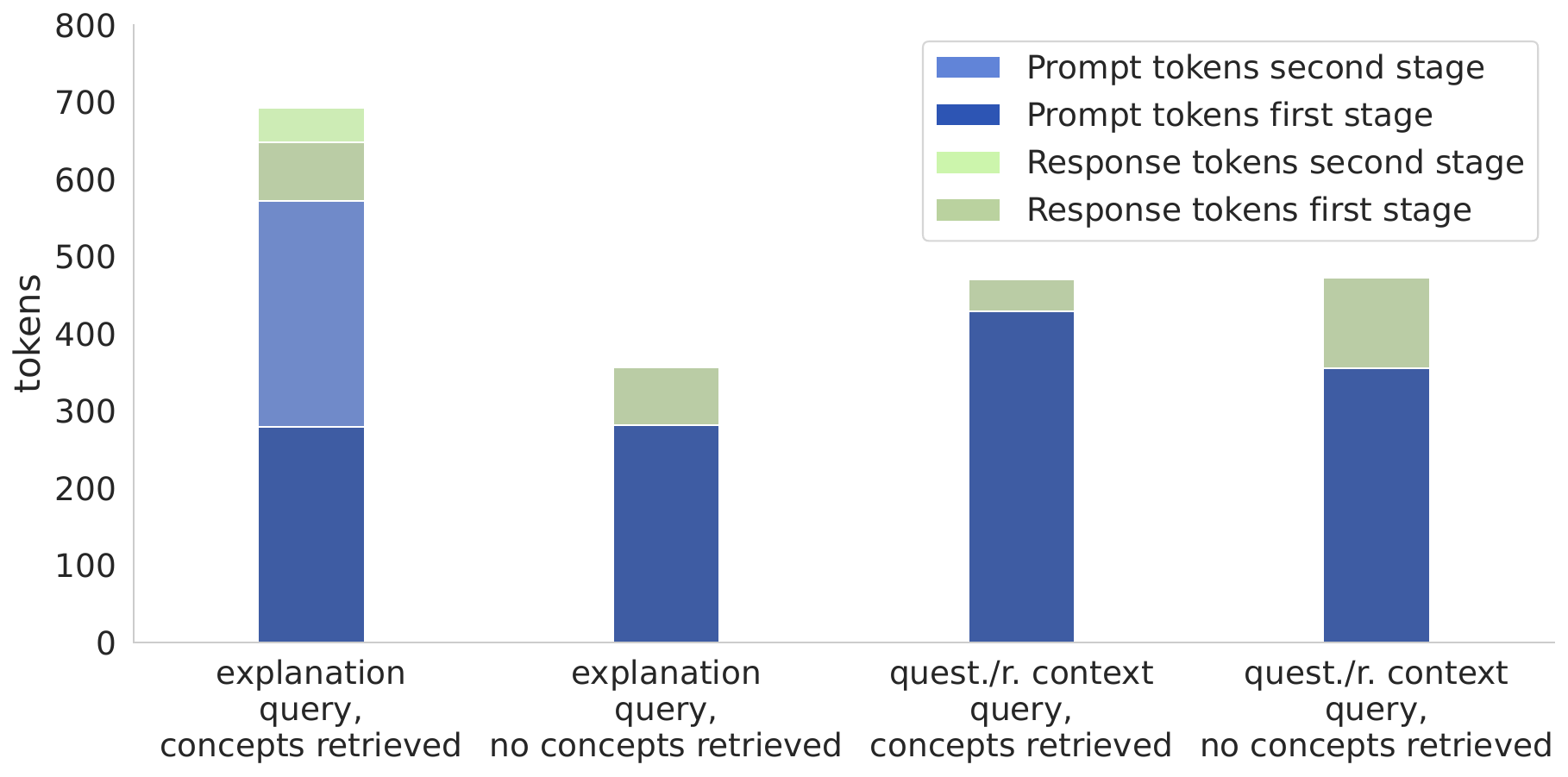}
\vspace{-0.5em}
\caption{Processed tokens in the prompt and response.}
\vspace{-0.8em}
\label{fig:tokens}
\end{figure}

Instead of comparing the inference speed, which depends on the LLM model size and available computing resources, we report the prompt and response token size (Fig. \ref{fig:tokens}), which proportionally influences the inference speed. In our case, we have used the \textit{OpenAI API}, achieving a mean inference time for the whole pipeline of 3.88 seconds per explanation. In Fig. \ref{fig:tokens}, we report separately experiments for the two-stage architecture, that is, the \textit{explanation query} architecture, and the one-stage architectures, which are the \textit{question query} and \textit{robot context query}. We also separate experiments where no memory concepts have been retrieved versus when at least one concept has been retrieved. The rationale behind this separation is that for the two-stage architecture, the second stage is not executed if no concepts are retrieved, forwarding the initial full explanation. We segment results for the token lengths among stages for the \textit{explanation query} architecture.

\section{DISCUSSION}
\label{sec:discussion}

Before the results' analysis, we manually inspected 10\% of the generated explanations across ablations, test cases and topics. This confirmed that the generated explanations were truthful and adequately addressed the root cause. This consistency is expected due to curated \textit{robot context} provided to the LLM. Our primary focus in this section is therefore analysing the adaptation of the explanation’s level of detail.

\subsection{Retrieved concepts}
When comparing the possible architectures in Table \ref{tab:results},  the best results—higher precision and recall and lower false positive rate—are obtained for the \textit{explanation query} setting, followed by the \textit{robot context query} and then \textit{question query}. These results are expected since the semantic similarity between the previously known concepts and the query will be higher for the explanation that is to be delivered, followed by the \textit{robot context} that is used to generate that explanation and, finally, by the user question. An illustrative example of a potential low similarity in the \textit{question query} is the following: the same question, e.g., \textit{``Why didn't you pick the apple?''}, can lead to different explanations because either the apple was not reachable or not seen.

Moreover, results in Table \ref{tab:results} report a general trend of higher precision, lower recall and lower false positive rates as the retrieval is more strict, moving towards higher $\tau_{retr}$ thresholds. This precision-recall trade-off needs to be assessed upon each application and tuned via $\tau_{retr}$ to balance the cases where relevant concepts are missed versus when incorrect concepts are introduced. For the \textit{unknown} topics, it is important to validate that the false positive rate is significantly low, which the ablations with $\tau_{retr} = high$ clearly achieve, as well as the \textit{explanation query} with $\tau_{retr} = mid$.

\subsection{Length reduction}

When comparing the discrepancy within the \textit{topic type} for the same \textit{elapsed time}, the length reduction difference between the \textit{known} and \textit{unknown} settings should be as high as possible, indicating that the system effectively reduces detail only when the user has related knowledge. Moreover, the \textit{unknown} topics should have a very low length reduction since there is no related knowledge, and thus, no adaptation of the level of detail should take place. As expected, the ablations with a $\tau_{retr} = zero$ do not fulfil these requirements because all concepts in the memory, including completely irrelevant ones, are retrieved and considered relevant. The \textit{question query} ablations moderately fulfil the requirements, but only for $\tau_{retr} = high$, which limits the length reduction for the \textit{known} topics. The \textit{robot context query} ablations present a wider difference between \textit{known} and \textit{unknown} \textit{topic types} due to closer similarity between the query and memory concepts that lead to a more accurate retrieval. The \textit{explanation query} ablation delivers the broadest difference between the \textit{known} and \textit{unknown} topics and a remarkably low length reduction for \textit{unknown} topics, with $\tau_{retr} = mid$ showing the best tradeoff.

These outcomes align with the retrieval results since a less accurate retrieval due to a lower semantic similarity directly influences the target level of detail, as presented in Sec. \ref{sec:approach}.

When comparing the \textit{1 week} and \textit{1 month} \textit{elapsed time} settings for the \textit{known} \textit{topic type}, the length reduction is higher for the \textit{1 week} case, as the concepts have a higher probability to be remembered and the level of detail is further reduced when compared to the \textit{1 month} case, where the probability that users retain the knowledge is lower.

\subsection{Token lengths}

Although the \textit{explanation query} ablation produces better results, an expected disadvantage is that the two-stage LLM calls can lead to a higher inference time. In Fig. \ref{fig:tokens}, we observe how the token length is effectively higher for \textit{explanation query} in the cases where concepts are retrieved, but shorter when there are no retrieved memories. In the latter, the second stage is skipped, forwarding the full initial explanation without processing the \textit{user context}, as explained in Sec. \ref{sec:approach}. Therefore, depending on the frequency of users asking about known vs. unknown topics, the two-stage approach can run faster or slower than the one-stage one.

\subsection{Future Work}

In this contribution, we have technically evaluated the proposed framework through 3 architecture implementations. Future work should study the effect of such personalised explanations on users due to an adequate level of detail. We are planning a longitudinal study where only part of the participants will receive personalised explanations, to measure if such adaptation improves the usage and usability without decreasing the understanding of the system.

Moreover, future research should investigate how to populate the memory with other multimodal sources, such as graphical user interfaces or observational systems. This way, facts that the user has experienced can update the estimated user knowledge in the memory representation on top of verbal explanations.
For this purpose, a new metadata field could be added to encapsulate the initial probability of being known, populated by attention models or adapted depending on the multimodal source.
This approach could require evaluating the differences between the estimated user memory and the concepts that users have actually learned.

\section{CONCLUSIONS}
\label{sec:conclusions}

In this work, we introduced a framework for generating personalised explanations. By integrating a user memory that provides relevant user context to an LLM, the system can dynamically adjust the explanations' level of detail. We compared three architectures and evaluated the levels of correct information retrieval and length reduction. Results indicate that the proposed framework effectively leverages the user's previous knowledge, especially in the two-stage architecture. Future research should evaluate in a user study the impact of the personalised explanations on users.

\section*{ACKNOWLEDGMENT}

This work has been partially supported by the Horizon Europe Marie Skłodowska-Curie grant agreement N. 101072488 (TRAIL), the Horizon Europe grant agreement N. 10107025 (CoreSense) and the Horizon 2020 grant agreement N. 857188 (SAFE-LY-PHARAON).

\bibliographystyle{abbrv}
\bibliography{refs}

\end{document}